\documentclass{article}

\usepackage[preprint, nonatbib]{neurips_2020}

\usepackage[utf8]{inputenc} 
\usepackage[T1]{fontenc}    
\usepackage{hyperref}       
\usepackage{url}            
\usepackage{booktabs}       
\usepackage{amsfonts}       
\usepackage{nicefrac}       
\usepackage{microtype}      
\usepackage{graphicx}
\usepackage{amsmath,amssymb}
\usepackage{epstopdf}
\DeclareMathOperator{\E}{\mathbb{E}}
\usepackage[square,numbers]{natbib}

\title{ICAM: Interpretable Classification via Disentangled Representations and Feature Attribution Mapping }

\author{%
  Cher Bass\\
  BME  \\
  King's College London\\
     \And
   Mariana da Silva\\
     BME  \\
  King's College London\\
   \And
   Carole Sudre\\
     BME  \\
  King's College London\\
   \And
   Petru-Daniel Tudosiu\\
     BME  \\
  King's College London\\
   \And
   Stephen M. Smith\\
   FMRIB\\
  University of Oxford \\
   \And
   Emma C. Robinson\\
     BME  \\
  King's College London\\
}

\begin{document}

\maketitle
\begin{abstract}
Feature attribution (FA), or the assignment of class-relevance to different locations in an image, is important for many classification problems but is particularly crucial within the neuroscience domain, where accurate mechanistic models of behaviours, or disease, require knowledge of all features discriminative of a trait. At the same time, predicting class relevance from brain images is challenging as phenotypes are typically heterogeneous, and changes occur against a background of significant natural variation.  Here, we present a novel framework for creating class specific FA maps through image-to-image translation. We propose the use of a VAE-GAN to explicitly disentangle class relevance from background features for improved interpretability properties, which results in meaningful FA maps. We validate our method on 2D and 3D brain image datasets of dementia (ADNI dataset), ageing (UK Biobank), and (simulated) lesion detection. We show that FA maps generated by our method outperform baseline FA methods when validated against ground truth. More significantly, our approach is the first to use latent space sampling to support exploration of phenotype variation. Our code will be available online at \url{https://github.com/CherBass/ICAM}.

\end{abstract}

\section{Introduction}

Brain images present a significant resource in the development of mechanistic models of behaviour and neurological/psychiatric disease as they reflect measurable neuroanatomical traits that are heritable, present in unaffected siblings and detectable prior to disease onset \cite{cullen2019polygenic}. Nevertheless, for complex disorders, features of disease remain subtle, variable \cite{iqbal2005subgroups,ross2006neurobiology} and occur against a back-drop of significant natural variation in shape and appearance \cite{glasser2016multi,kong2019spatial}.

Traditional approaches for brain image analysis compare data in a global average template space, estimated via smooth (and ideally diffeomorphic) deformations \cite{avants2010optimal,dalca2019learning, dalca2019unsupervised, evans2012brain,  glasser2016multi,robinson2018multimodal}. This, however, typically ignores cortical heterogeneity and may smooth out sources of variation \cite{dalca2019learning,glasser2016multi} in ways which limit interpretation \cite{bijsterbosch2018relationship}. Tools are still required to distinguish between features of population variability and specific discriminative phenotypic features.


Deep learning is state-of-the-art for many image processing tasks \cite{goodfellow2016deep} and has shown strong promise for brain imaging applications such as healthy tissue and lesion segmentation \cite{chen2018voxresnet,de2015deep,kamnitsas2017ensembles,rajchl2018neuronet}. However, there is growing need for greater accountability of networks, especially within the medical domain. Several approaches for feature attribution \cite{bach2015pixel,lundberg2017unified,montavon2016deeptaylor,selvaraju2017grad,sundararajan2017axiomatic} have been proposed which return the most important or salient features for a prediction, after training a network for classification. However, applying a method post-hoc instead of explicitly training an interpretable model has shown to be insufficient at detecting all discriminative features of a class, especially in medical imaging \cite{baumgartner2018visual}. Recently, approaches have been proposed which use generative models to translate images from one class to another \cite{isola2017image, zhu2017unpaired}, and in \citet{baumgartner2018visual} this was adapted to create a difference map between Alzheimer's (AD) and Mild Cognitive Impairment (MCI) subjects. While it was able to detect more salient features in comparison to previous methods, it was still unable to identify much of the variability between the AD and MCI subjects. Other restrictions of this approach are that it assumes knowledge of label classes at test time, and that it does not have a latent space that can be sampled. This limits the interpretation and generates a deterministic output in test time.

\begin{figure}[!t]
  \centering
\makebox[\linewidth]{
	\includegraphics[width=1.0\textwidth]{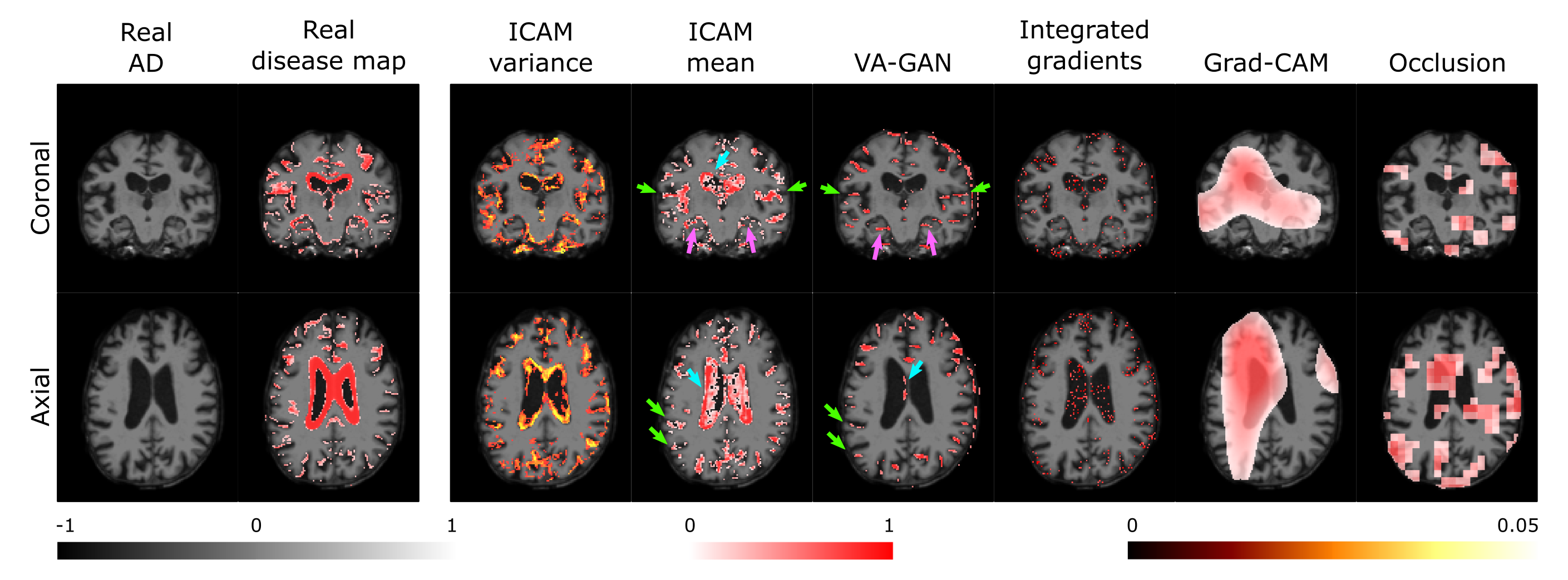}}
\caption{ADNI comparisons of Feature Attribution (FA) maps. ICAM is the first known method able to generate variance and mean FA maps in test time, and shows good detection of the ventricles (blue arrows), cortex (green arrows), and hippocampus (pink arrows) when compared with the ground truth disease map. Baseline methods perform sub-optimally in comparison, with VA-GAN generating the second best FA maps.}
\label{fig:adni_comparison}
\end{figure}

In this paper we aim to improve on the current feature attribution methods by developing a more interpretable model, and thus more meaningful feature attribution (FA) maps. We propose ICAM (Interpretable Classification via disentangled representations and feature Attribution Mapping) a framework which builds on approaches for image-to-image translation \cite{lee2019drit} to learn feature attribution by disentangling class-relevant \textit{attributes} (attr) from class-irrelevant \textit{content}. Sharp reconstructions are learnt through use of a Variational Autoencoder (VAE) with a discriminator loss on the decoder (Generative Adversarial Network, GAN). This not only allows classification and generation of an attribution map from the latent space, but also a more interpretable latent space that can visualise differences between and within classes. By sampling the latent space in test time to generate a FA map, we demonstrate its ability to detect meaningful brain variation in 3D brain Magnetic Resonance Imaging (MRI).

In particular the specific contributions of the method are as follows:
\begin{itemize}
\item[1] We describe the first framework to implement a translation VAE-GAN network for simultaneous classification and feature attribution, through use of a shared attribute latent space with a classification layer, which supports rejection sampling and improved class disentanglement, relative to previous methods \cite{lee2019drit}. 
\item[2] This supports exploration of phenotypic variation in brain structure through latent space visualisation of the space of class-related variability, including the study of the mean and variance of feature attribution map generation (see example in Fig.~\ref{fig:adni_comparison}).
\item[3] We demonstrate the power and versatility of ICAM using extensive qualitative and quantitative validation on 3 datasets including; Human Connectome Project (HCP) with lesion simulations, ADNI, and UK Biobank datasets. In addition, our code, which will be released on GitHub, extends to multi-class classification and regression tasks. 
\end{itemize}

\section{Related works}

\subsection{Feature attribution methods}
The most commonly used approach for feature attribution follows the training of a classification network with importance or saliency mapping (Table \ref{table:visual_attribution_methods}, rows 1-3). These typically analyse the gradients or activations of the network, with respect to a given input image, and include approaches such as Gradient-weighted Class Activation Mapping (Grad-CAM) \cite{selvaraju2017grad}, SHAP \cite{lundberg2017unified}, DeepTaylor \cite{montavon2016deeptaylor}, integrated gradients \cite{sundararajan2017axiomatic}, and Layer-wise backpropagation (LRP) \cite{bach2015pixel}. In contrast, perturbation methods such as occlusion \cite{zeiler2014visualizing} change or remove parts of the input image to generate heatmaps, by evaluating its effect on the classification prediction. Most of these methods, however, provide coarse and low resolution attribution maps. More importantly, they depend on a network trained prior to applying a particular FA method, and will perform sub-optimally if the network learnt to focus on only the most salient features relevant to a class (e.g. focus on a dog's face for natural image prediction, but not on its tail or other distinctive features); this is common with classification networks.

\begin{table}
  \footnotesize
  \caption{Comparison of baseline methods.}
  \label{table:visual_attribution_methods}
  \centering
  \begin{tabular}{llllll}
    \toprule
    \cmidrule(r){1-2}
    Method & post-hoc & classification & generative model & cyclic &variance analysis\\
    \midrule
    Grad-CAM \cite{selvaraju2017grad}& \checkmark  &  \checkmark&  & N/A  &  \\
    Integrated gradients \cite{sundararajan2017axiomatic} & \checkmark  & \checkmark  &  &N/A &\\
    Occlusion \cite{zeiler2014visualizing} & \checkmark  & \checkmark &  & N/A & \\
    DRIT \cite{lee2019drit} & &  & \checkmark    & \checkmark & \\
    VA-GAN \cite{baumgartner2018visual} & &  & \checkmark    & & \\
    ICAM (our method) & & \checkmark & \checkmark & \checkmark & \checkmark\\
    \bottomrule
  \end{tabular}
\end{table}

\subsection{Generative models}
\begin{figure}[!b]
  \centering
\makebox[\linewidth]{
	\includegraphics[width=0.8\textwidth]{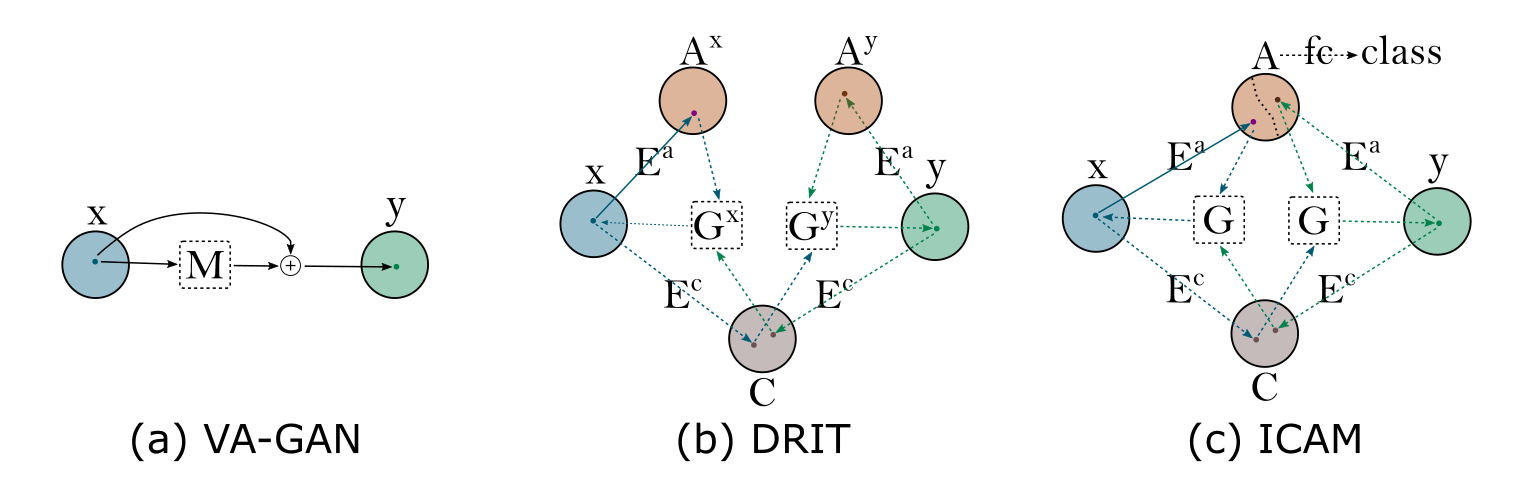}}
\caption{Comparison of domain mapping methods.}
\label{fig:I2I_translation}
\end{figure}

An alternative approach is the use of generative models, specifically GANs and VAEs, where a common application, image-to-image translation, has been used successfully in many different domains \cite{isola2017image,zhu2017unpaired, huang2018multimodal,liu2017unsupervised, jha2018disentangling, lee2019drit}, including medical imaging \cite{baumgartner2018visual, bass2019image, baur2018deep, costa2017end}. Of these, \citet{lee2019drit}, in  particular, developed a domain translation network called DRIT (Fig.~\ref{fig:I2I_translation}b), which constraints features specific to a class, through encoding separate class-relevant (attribute) and class-irrelevant (content) latent spaces, and employing a discriminator (Table \ref{table:visual_attribution_methods}, row 4).

Separately, \citet{baumgartner2018visual},  developed a conditional GAN-based approach, called VA-GAN, for feature attribution, using domain translation between 3D MRI of brains with AD and MCI (Table \ref{table:visual_attribution_methods}, row 5). In this work, a mapping $M$ is learnt, which translates an AD input image towards the MCI class (Fig.~\ref{fig:I2I_translation}a), resulting in sharp reconstructions and realistic difference maps that overlap with ground truth outcomes, where available. One constraint of VA-GAN, however,  is that it requires image class labels to be known a \textit{priori}. And, in the absence of a latent space, it can only produce a single deterministic output for an input image, which limits the interpretation in comparison to methods with a latent space.

In this paper, we therefore extend the intuitions of these models to create one framework which allows simultaneous classification and feature attribution, using a more interpretable model. In particular, we use a VAE-GAN to encode a class-relevant attribute latent space which is shared between classes, and thus allows classification and the visualisation of differences between and within classes (Fig.~\ref{fig:I2I_translation}c). Table~\ref{table:visual_attribution_methods} summarises the advantages of ICAM relative to other feature importance, saliency, and generative visualisation methods.

\section{Methods}
\subsection{Method overview} 
The goal of our framework is to learn a classifier that encodes inputs of different classes into a separable latent space, and a generator that synthesizes FA maps with all class-relevant salient features. We use a VAE-GAN with an adversarial content latent space, which learns about class-irrelevant information, and an attribute latent space, which learns about class-relevant information. The overview of the framework is shown in (Fig.~\ref{fig:network_translation}). The approach is comprised of the following components:

\begin{figure}[!t]
 \centering
\makebox[\linewidth]{
	\includegraphics[width=0.9\textwidth]{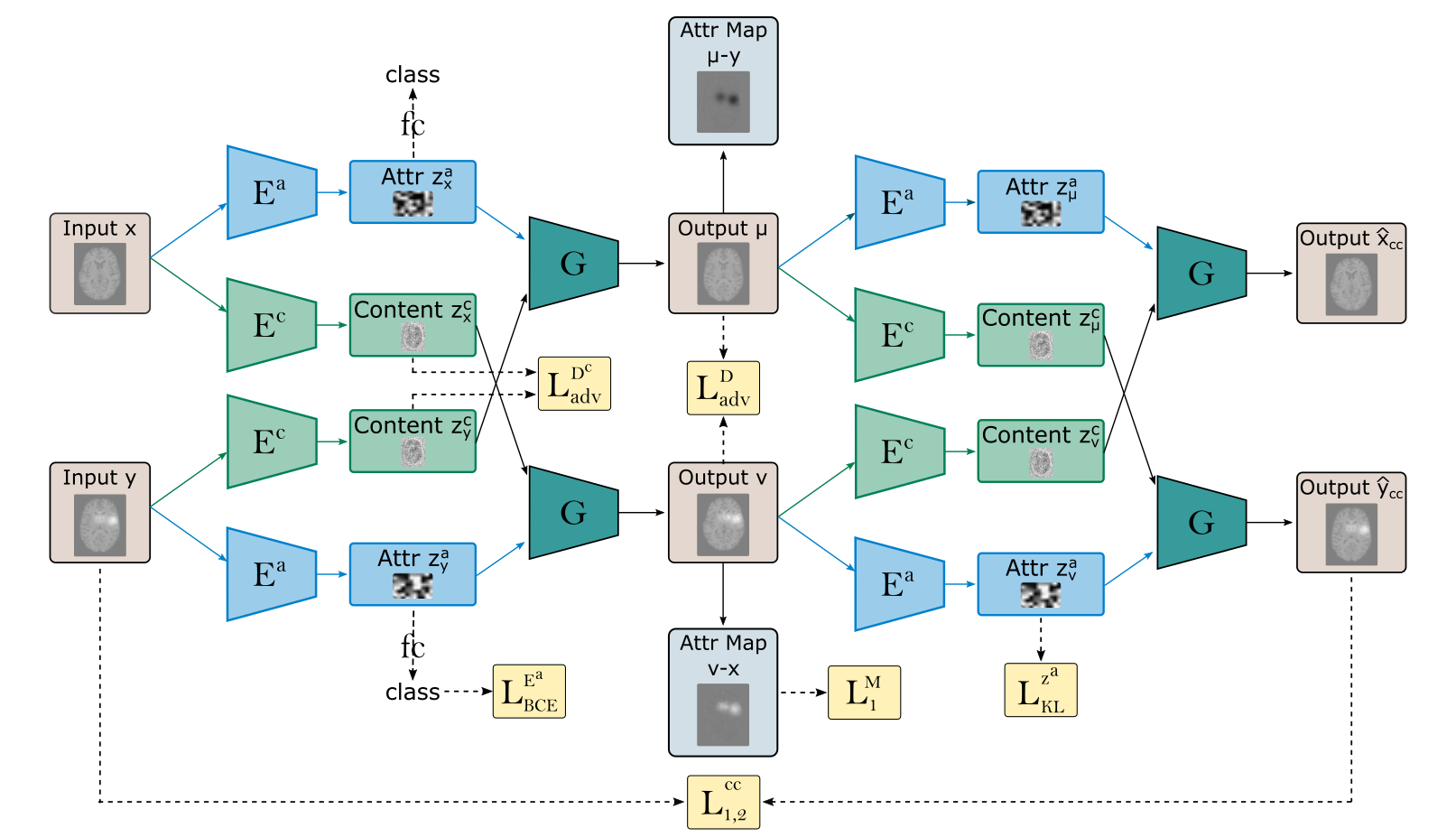}}
\caption{Overview of method. An example of how ICAM performs classification with attribute map generation for 2 given input images x and y (of class 0 [brain slice without lesions] and 1 [brain slice with blob-like lesions], respectively). Note that not all losses are plotted (see Equation \ref{eq:objective} for full objective).}
\label{fig:network_translation}
\end{figure}

A \textbf{content encoder} $\{E^c\}$ encodes class-irrelevant  information using a shared content latent space $\{z_x^c, z_y^c \in C\}$, via the application of a \textbf{content discriminator} $\{D^c\}$, whose goal is to discriminate between the classes or domains. For input images $\{x, y\}$ of classes $\{c_x\}$ and $\{c_y\}$, respectively, the goal of the content encoder $\{E^c\}$ is to fool the discriminator to classify an image incorrectly, and to make the content latent space appear the same, regardless of the class $(E^c: x \rightarrow z_x^c), (E^c: y \rightarrow z_y^c)$. An \textbf{attribute encoder} $\{E^a\}$ learns all relevant class information, and classifies between domains $(E^a: x \rightarrow z_x^a \rightarrow c_x), (E^a: y \rightarrow z_y^a \rightarrow c_y)$ using a fully connected/ dense layer that is applied to the shared attribute (class) latent space $\{z_x^a, z_y^a \in A\}$. The \textbf{generator} $\{G\}$ learns to synthesise an image conditioned on both the content and attribute latent spaces $(G: \{z_x^c, z_x^a\} \rightarrow \hat{x}), (G: \{z_y^c, z_y^a\} \rightarrow \hat{y})$, as well as to translate between these domains, by swapping the content latent space; this is possible since the content latent space is class invariant $(G: \{z_y^c, z_x^a\} \rightarrow \mu), (G: \{z_x^c, z_y^a\} \rightarrow v)$. Translating the domains enables the visualization of differences between the two classes, using a \textbf{feature attribution map} $(\{M_{x} = v-x\}, \{M_{y} = \mu-y\})$. Finally, the \textbf{domain discriminator} $\{D\}$ learns to distinguish between fake and real images, and to classify the two domains, which gives a clearer training signal for the generator. 

Our network architectures uses 2D or 3D convolutional layers (kernel size = 3 or 4), ResNet layers including basic, down, and deconvolutional blocks with instance normalisation. A key feature of the network architecture is encoding the latent space as a 2D or 3D vector, instead of a 1D vector as is commonly seen in VAEs. For example, for a latent space size of 80, a 1D vector = [80], 2D vector = [8,10]. This allows the network to encode spatial, and shape related information in the latent space, which is important in brain imaging. A detailed diagram of our encoder-generator architecture (for a 3D input) is shown in section \ref{sec:architecture}.

\subsection{Content and attribute spaces} 
Our approach disentangles the two image domains into a shared content space $\{C\}$, and attribute space $\{A\}$. The content latent space aims to encode class irrelevant information: features common to both domains (e.g. the shape, location of structures, and folds of the brain). The attribute latent space aims to map the remaining domain-specific information onto shared latent space $\{A\}$, thus enabling classification using all salient features, and cross-domain translation. 

\paragraph{Content loss.}
To achieve domain disentanglement, we employ a common content encoder for both domains $(\{E_c : x \rightarrow C\}, \{E_c : y \rightarrow C\})$. The content latent space $\{C\}$ is fed into a content discriminator, $\{D^c\}$, which outputs the image class probability. The content discriminator $\{D^c\}$ aids the representation to be disentangled, by aiming to distinguish between domains (classes) of the encoded latent spaces $\{z_x^c\}$ and $\{z_y^c\}$. Inversely, the content encoder $\{E_c\}$ aims to encode a representation whose domain cannot be distinguished by the content discriminator, and thus forces the representation to be mapped to the same space $\{C\}$, similarly to \citet{lee2019drit}.

\noindent This class adversarial content loss can be expressed as:
\begin{equation}
  \begin{aligned}
L_{adv}^{D^c} = \E_{z_x^c}[\log D^c(E^c(x)) + \log(1-D^c(E^c(x)))] \\
+ \E_{z_y^c}[\log D^c(E^c(y)) + \log(1-D^c(E^c(y)))].
  \end{aligned}
\end{equation}

\noindent An L2 regularisation was added to prevent explosion of gradients, while Gaussian noise was added to the last layer of the content encoder to prevent vanishing of the content latent space.

\paragraph{Classification loss.}
Classification is performed through extending the attribute latent space using a fully connected layer with binary cross entropy loss $L_{BCE}^{E^a}$, to encourage the separation of the domains within the shared attribute latent space $\{A\}$.

\paragraph{VAE loss.}
We employ a latent variable model, where we place a Gaussian prior over the latent variables and train using variation inference, by applying the Kullback Leibler (KL) loss $L_{KL}^{z^a}$.



\paragraph{Latent regression loss.}
We impose an additional loss on the attribute latent space in order to encourage invertible mapping between the image and the latent space, which also aids the cyclic reconstruction. We sample an attribute latent vector $z^a_r$ from a Gaussian distribution, and attempt to reconstruct it:
\begin{equation}
L_1^{z^a} = \|E^a(G(E^c(x),z^a_r))-z^a_r\|_1. 
\end{equation}

\paragraph{Rejection sampling of the attribute latent space.}
Disentanglement is further encouraged through `rejection' sampling of the attribute latent space during training by checking the class of a randomly sampled vector using the attribute encoder's classification layer. Samples are rejected if they belong to the wrong class, which stabilises optimisation of translation by passing the generator samples of the expected class. This further allows translation (using a single image) in test time, by first encoding an input image (Fig.~\ref{fig:random_sampling}a), and then sampling from the space until the opposite class is sampled (Fig.~\ref{fig:random_sampling}b), by checking the random vector's class using the classifier. By sampling multiple times, we can also get mean and variance FA maps during test time (See an example in Fig.~\ref{fig:adni_comparison} and \ref{fig:biobank_comparison}), a visualisation approach previously not possible in other feature attribution methods, as they do not have a latent space with a classification layer.

\subsection{Generation and feature attribution}

\paragraph{Feature attribution (FA) map loss.}
To visualise differences between the translated images $\{v, \mu\}$ and the original images $\{x, y\}$, we use a feature attribution map $\{M\}$. This aims to retain only class-related differences between two images (or two locations in the attribute latent space) by subtracting the content from the translated output $(\{M_{x} = v-x\}, \{M_{y} = \mu-y\})$. Generation is regularised through an L1 loss ($L_1^{M} = \| M(\;)\|_1,$) which encourages $\{M\}$ to reflect a small feasible map, which leads to a realistic translated image. 

\paragraph{Domain loss.}
The domain loss combines a domain adversarial loss, $L^{D}_{adv}$, (to discriminate between real and fake images, and encourage realistic image generation) and a binary cross entropy loss, $L^{D}_{BCE}$ (to encourage generation of images of the expected class).

\paragraph{Reconstruction loss.}
To facilitate the generation, we apply an L1 and L2 loss to the reconstructed images $\{\hat{x}, \hat{y}\}$ ($L_1^{rec}$), and the cyclically reconstructed images $\{\hat{x_{cc}}, \hat{y_{cc}}\}$ ($L_1^{cc}$). The cycle consistency term also allows training with unpaired images.

\begin{equation}
L_{1,2}^{rec} = \E_{x,y}[\|G(E^c(x), E^a(x))-x\|_{1,2} + \|G(E^c(y), E^a(y))-y\|_{1,2}],
\end{equation}

\begin{equation}
L_{1,2}^{cc} = \E_{x,y}[\|G(E^c(v), E^a(\mu))-x\|_{1,2} + \|G(E^c(\mu), E^a(v))-y\|_{1,2}].
\end{equation}

This means the \textbf{full objective function}\footnote{see section \ref{sec:training} for $\lambda$ values during training.} of our network is:
\begin{equation}
  \begin{aligned}
\min_{G, E^c, E^a} \max_{D, D^c}  \lambda_{D^c} L_{adv}^{D^c} + \lambda_{D} L^{D}_{adv} + \lambda_{D_{BCE}} L^{D}_{BCE} + \lambda_{BCE} L_{BCE}^{E^a} + \lambda_{KL} L_{KL}^{z^a}\\ + \lambda_{M} L_1^{M}
+\lambda_{z^a} L_1^{z^a} + \lambda_{rec} (L_1^{rec} + L_1^{cc} + L_2^{rec} +  L_2^{cc}) .
  \end{aligned}
  \label{eq:objective}
\end{equation}

\section{Results}

We evaluate the performance of ICAM through studies on three datasets to perform 1) ablation studies on 2D simulations; 2) evaluation of the accuracy of the generated attribution maps (using ground truth disease conversion maps derived from the ADNI dataset); 3) exploration of the flexibility of the approach for investigating phenotypic variation (using healthy ageing data from UK Biobank). 

\subsection{Comparison methods and metrics}
We compared our proposed approach against a range of baselines in our experiments. For a fair comparison, we use the same training, validation and testing datasets. In particular, we compared against \textbf{Grad-CAM} \cite{selvaraju2017grad}, \textbf{Integrated gradients} \cite{sundararajan2017axiomatic}, \textbf{Occlusion} \cite{zeiler2014visualizing}, \textbf{VA-GAN} \cite{baumgartner2018visual}, and \textbf{DRIT++} \cite{lee2019drit}. We compare against 2 variations of the DRIT++ network, the original, $DRIT_{z_8}$, and with increased attribute latent space size ($DRIT_{z_{80}}$, size of 80 rather than 8), which is the same as ICAM. For further details on comparison methods, refer to section \ref{sec:comparison_methods}. We also compared against different variations of our network, \textbf{ICAM}, including: $ICAM_{DRIT}$, uses ICAM architecture, but all the same losses as in DRIT; $ICAM_{BCE}$, which adds the classification BCE loss to the attribute latent space and rejection sampling during training; $ICAM_{FA}$, which adds the FA map loss; and finally $ICAM$, adding the l2 loss and thus containing all described losses in this paper.

Networks are compared using accuracy score for classification, and normalised cross correlation (NCC) between the absolute values of the attribution maps and the ground truth masks (e.g. the lesion masks in the HCP lesion simulations, or the disease effect maps in ADNI). The positive NCC (+) compares the lesion mask to the attribution map when translating between class 0 (e.g. no lesions, or MCI) to 1 (e.g. lesions, or AD), and vice versa for the negative NCC (-). Values reported are the mean and standard deviation across the test subjects.

\subsection{HCP 2D ablation experiments}
Many neurological and psychiatric disorders show variability in presentation and symptoms, meaning that it is common to observe differences in the imaging phenotype. In our ablation experiments, we used the HCP dataset \cite{glasser2016multi, van2012human}, with T2 MRI data, for simulating lesions in MRI brain slices, to mimic this type of variability. We use the original data as class 0 (no lesions), and then create cortical 'lesions' which appear in the 'disease' class 1, with different frequencies. During training, we take 2D axial slices from the centre of the brain (to which all regions are constrained by design), so that we can compare against DRIT, which is a 2D network. For further details on the HCP dataset, see section \ref{sec:HCP}.

Ablation results are displayed in Table \ref{table:ablation_study}, where we show that $ICAM_{DRIT}$ network (i.e. ICAM architecture, with the same losses as DRIT) performs similarly to DRIT, whilst having a much more compact architecture. Due to memory constraints, only ICAM supports extension to 3D (with 1.8M rather than 2.6M trainable parameters, making the 3D encoder-decoder network of DRIT $1.4\times$  bigger). Furthermore, we show that the addition of different components of the network improves performance further (see NCC scores, Table \ref{table:ablation_study}, rows 3-6), with the best overall performance achieved by the full ICAM network (Table \ref{table:ablation_study}, row 6). In addition, we observe that qualitatively ICAM performs better when interpolating between and within classes (Fig.~\ref{fig:interpolation}). ICAM FA maps appear to be smoothly changing when interpolating between the lesion and no lesion class, while in DRIT the FA maps are similar across the interpolation, suggesting that some of the class information is encoded in the content latent space. Further, we observe that ICAM is able to both add and remove lesions simultaneously while interpolating within the lesion class, whereas DRIT is only able to remove lesions. Overall, this indicates that ICAM has achieved better separation in the attribute latent space, and is further supported by the tSNE plots within class 1 (Fig.~\ref{fig:tsne}). 

\begin{table}
\small
  \caption{Ablation experiments using the 2D HCP dataset with lesion simulations.}
  \label{table:ablation_study}
  \centering
  \begin{tabular}{llll}
    \toprule
    \cmidrule(r){1-2}
    Network     & Accuracy & NCC (-) & NCC (+)   \\
    \midrule
    $DRIT_{z_8}$ & N/A & 0.346 $\pm$ 0.080 & 0.243 $\pm$ 0.050     \\
    $DRIT_{z_{80}}$ & N/A  & 0.380 $\pm$ 0.081 & 0.272 $\pm$ 0.068  \\
    $ICAM_{DRIT}$ & N/A  & 0.385 $\pm$ 0.094 & 0.265 $\pm$ 0.062  \\
    $ICAM_{BCE}$ & 0.899  & 0.316 $\pm$ 0.102 & 0.325 $\pm$ 0.100  \\
    $ICAM_{FA}$ & 0.900   & 0.353 $\pm$ 0.104 &  \bf 0.333 $\pm$ 0.082     \\
    $ICAM$ & \bf 0.950  & \bf 0.435 $\pm$ 0.092 & 0.332 $\pm$ 0.098    \\
    \bottomrule
  \end{tabular}
\end{table}

\subsection{ADNI experiments: Ground-truth evaluation of feature attribution maps} 
We use the longitudinal ADNI dataset \cite{jack2008alzheimer}, with T1 MRI data, as in \citet{baumgartner2018visual}, to evaluate feature attribution maps, generated by ICAM, and other baseline methods. DRIT \cite{lee2019drit} was not used as the network design cannot scale to 3D due to memory constraints. ADNI contains paired examples for which images that are acquired before and after conversion to an AD state; where the intermediate state between healthy cognition and AD is known as MCI. We used these subjects in our validation and test sets, to calculate the ground truth disease map (i.e. AD-specific brain atrophy patterns) for each subject, which we then compared against the FA maps of each method to compute the NCC score. See section \ref{sec:ADNI} for further details on the dataset. 

We found that ICAM outperforms VA-GAN when comparing the NCC metric (Table~\ref{table:adni}), and that they both perform better than occlusion, integrated gradients and Grad-CAM. The FA maps generated by the VA-GAN in our comparisons differ to the original results of VA-GAN \cite{baumgartner2018visual}, and these differences might be accounted by a more strict data selection process (we picked images acquired with 3T only, and did not combine with 1.5T acquired images as in \citet{baumgartner2018visual}), which resulted in a smaller training size (5778 vs 931 subjects for training in VA-GAN and this work, respectively), and small differences in the pre-processing. In our experiments, we observe that ICAM is able to detect most of the real disease effects in ventricles, cortex, and hippocampus, but that VA-GAN only detects some of these differences (Fig.~\ref{fig:adni_comparison}, and Fig.~\ref{fig:adni_extra_comparisons}). Both ICAM and VA-GAN generate higher resolution, and more interpretable FA maps in comparison to occlusion, integrated gradients and Grad-CAM, suggesting that using generative models, instead of a simple classification CNN, might be a better approach for detecting more discriminative features, and phenotypic variability.

\begin{table}[ht]
\small
  \caption{ADNI experiments.}
  \label{table:adni}
  \centering
  \begin{tabular}{llll}
    \toprule
    \cmidrule(r){1-2}
    Network & NCC (-) & NCC (+)  \\
    \midrule
    Occlusion \cite{zeiler2014visualizing} & $0.360 \pm 0.037$ &   $0.354 \pm 0.057 $  \\
    Grad-CAM \cite{selvaraju2017grad} & $0.321 \pm 0.059$ & $0.461 \pm 0.086 $    \\
    Integrated gradients \cite{sundararajan2017axiomatic} & $0.378 \pm 0.064 $ &  $0.404 \pm 0.059 $   \\
    VA-GAN \cite{baumgartner2018visual} & $0.653\pm 0.142$  &  N/A      \\
    ICAM & \bf 0.683 $\pm$ 0.097 & \bf 0.652 $\pm$ 0.083    \\
    \bottomrule
  \end{tabular}
\end{table}


\subsection{Biobank experiments} 

We used the UK Biobank \cite{alfaro2018image, miller2016multimodal}, with T1 MRI data, a collection of brain imaging data of mostly healthy subjects between the ages of 44-80 years old, to study phenotypic variation that occurs during ageing. To use this dataset for classification, we split the data into 2 classes, of young (class 0, 45-60 years) and old subjects (class 1, 70-80). For further details on the dataset see section \ref{sec:biobank}.

\begin{figure}[!b]
  \centering
\makebox[\linewidth]{
	\includegraphics[width=0.9\textwidth]{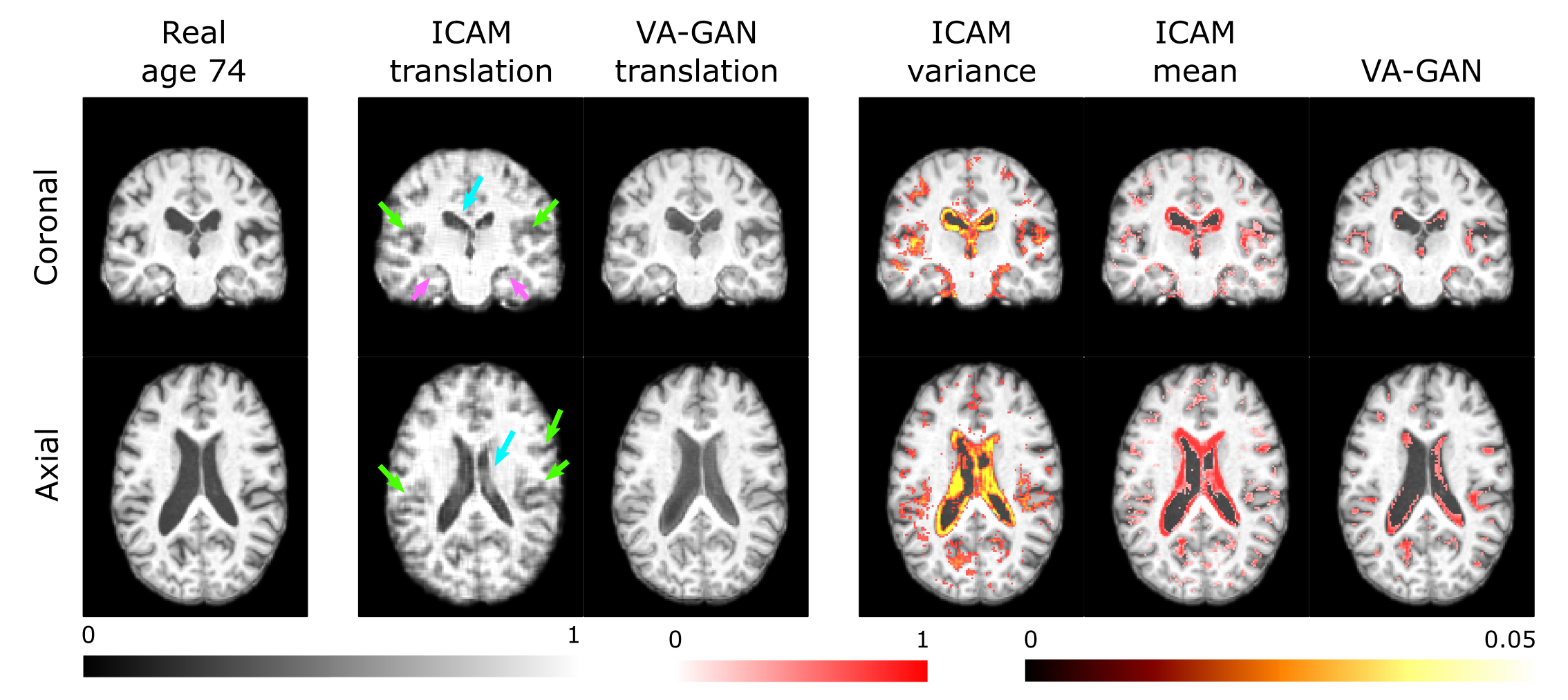}}
\caption{Biobank comparisons: modelling healthy ageing (translation of an old subject to young). We show that ICAM has visibly better detection in the hippocampus (pink arrows), ventricles (blue arrows), and cortex (green arrows). Further, ICAM is able to change the shape of different brain regions, whereas VA-GAN is only able to make minor adjustments in pixel intensities.}
\label{fig:biobank_comparison}
\end{figure}

We achieved 0.943 accuracy on the test dataset with ICAM on age classification, but cannot compare to VA-GAN, as it cannot perform class prediction. We compare the feature attribution maps and translated images against VA-GAN (Fig.~\ref{fig:biobank_comparison}, and Fig.~\ref{fig:biobank_very_old_young}). While we do not have ground truth maps for a comparison in Biobank, it is commonly described within the neuroscience domain that hippocampal atrophy \cite{o2016cognitive}, decrease in cortical thickness, and the enlargement of the ventricles are often observed with ageing, and are indicative of neurodegeneration \cite{guo2017mri, pacheco2015greater}. We looked for this phenotypic variation in our qualitative analysis. For ICAM we plot  variance, and mean FA maps (computed through multiple sampling of the attribute latent space in test time); for VA-GAN we show only FA as it can only generate a single deterministic output. We found that VA-GAN was able to detect some variation in the ventricles and cortex, but that ICAM was able to detect more variability in the ventricles (blue arrows), cortex (green arrows), and in the hippocampus (purple arrow). Interestingly, the variance map detects higher variability in the ventricles and the cortex, which may reflect heterogeneity in the ageing process. We also note that while ICAM is able to change the shape of different brain regions in translation (e.g. shrinking the ventricles, or enlarging the cortical folds), VA-GAN is only able to change the pixel intensities, and does not appear to change the shape. This is a significant advantage for ICAM compared to previous methods, as shape is an important phenotype in many other medical datasets \cite{garcia2018dynamic, kapellou2006abnormal}, and thus is likely to generalise well. 

To demonstrate the added value of the components introduced in ICAM (in comparison to DRIT), we also compared against a basic version of our network, $ICAM_{DRIT}$, and show that (1) the full version of ICAM, but not $ICAM_{DRIT}$, is able to separate the old and young subjects in the latent space (Fig.~\ref{fig:tsne_biobank}), and is thus (2) able to interpolate between subjects in the latent space (Fig.~\ref{fig:biobank_interpolation}).

\section{Conclusion}

In this work we developed a novel framework for classification with feature attribution. We demonstrate that our method achieves better performance on the HCP with lesion simulation, ADNI and UK Biobank datasets, compared to previous work. 
To our knowledge, it is the only approach able to generate FA maps directly from the attribute (class-relevant) and content (class-irrelevant) latent spaces, and its highly interpretable latent space allows detailed analysis of phenotypic variability, by studying the variance and mean FA maps. Finally, our code supports generalisation between 2D and 3D image spaces, and extensions to multi-class classification and regression.

\section*{Broader impact}

There is growing evidence that deep learning tools have the potential to improve the speed of review of medical images and that their sensitivity to complex high-dimensional textures can (in some cases) improve their efficacy relative to radiographers \cite{hosny2018artificial}. A recent study by Google DeepMind \cite{mckinney2020international} suggested that deep learning systems could perform the role of a second-reader of breast cancer screenings to improve the precision of diagnosis relative to a single-expert (which is standard clinical practice within the US).

For brain disorders the opportunities and challenges for AI are more significant since, the features of the disease are commonly subtle, presentations highly variable (creating greater challenges for physicians), and the datasets are much smaller in size in comparison to natural image tasks. The additional pitfalls that are common in deep learning algorithms \cite{vayena2018machine}, including the so called `black box' problem where it is unknown why a certain prediction is made, lead to further uncertainly and mistrust for clinicians when making decisions based on the results of these models.

We developed a novel framework to address this problem by deriving a disease map, directly from a class prediction space, which highlights all class relevant features in an image. We demonstrate on a theoretical level, that the development of more medically interpretable models is feasible, rather than developing a diagnostic tool to be used in the clinic. However, in principle, these types of maps may be used by physicians as an additional source of data in addition to mental exams, physiological tests and their own judgement to support diagnosis of complex conditions such as Alzheimer's, autism, and schizophrenia. This may have significant societal impact as early diagnosis can improve the effectiveness of interventional treatment. Further, our model, ICAM, presents a specific advantage as it provides a `probability' of belonging to a class alongside with a visualisation supporting better understanding of the phenotypic variation of these diseases, which may improve mechanistic or prognostic modelling of these diseases.

There remain ethical challenges as errors in prediction could influence clinicians towards wrong diagnoses and incorrect treatment which could have very serious consequences. Further studies have shown clear racial differences in brain structure \cite{shi2017using, tang2018brain} which if not sampled correctly could lead to bias in the model and greater uncertainty for ethnic minorities \cite{safdar2020ethical}. These challenges would need to be addressed before any consideration of clinical translation.  Clearly, the uncertainties in the model should be transparently conveyed to any end user, and in this respect the advantages of ICAM relative to its predecessors are plain to see.

\medskip
\small
\bibliographystyle{abbrvnat}
\bibliography{neurips_2020}

\appendix

\clearpage
\section{Methods} 
\subsection{Network Architecture}  \label{sec:architecture}

\begin{figure}[!ht]
  \centering
\makebox[\linewidth]{
	\includegraphics[width=1.0\textwidth]{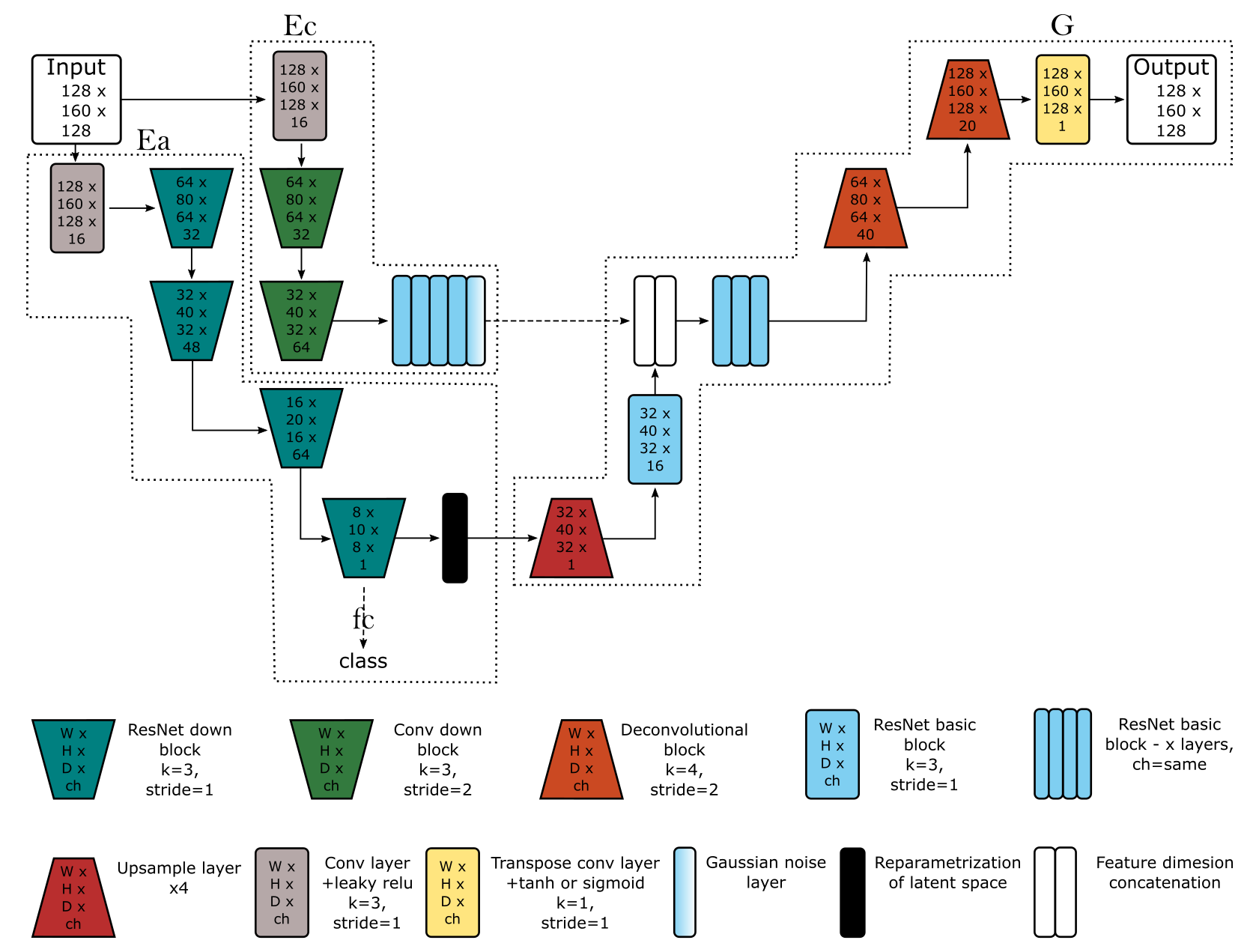}}
\caption{Network Architecture for 3D inputs.}
\label{fig:network_architecture}
\end{figure}

Our encoder-decoder architecture for a 3D input is shown is Fig.~\ref{fig:network_architecture}. The architecture for a 2D input is the same, only using 2D convolutions and a 2D attribute space. Here, an input image is encoded using 2 shared networks, the attribute encoder $E_a$, and the content encoder $E_c$, and then is reconstructed or translated (to another class) using the generator, $G$. 

The key components of the attribute encoder include using down ResNet blocks (with average pooling, and leaky ReLU activation) for encoding the input image into a relatively large 3D latent space of size $8 \times 10 \times 8$ (in the 3D case), as opposed to a 1D vector, which is commonly seen in Variational Autoencoders (VAEs). We also added a fully connected layer to the attribute latent space to enable classification. In early development, we found that using a 1D vector in the latent space was insufficient for encoding the required class information for brain imaging, and observed that some class information was instead encoded in the content encoder, which is meant to be class invariant. Using a sufficiently large 2D or 3D vector (depending on the input) helped with addressing this problem. 

The goal of the content encoder is to encode a class-irrelevant space, which allows translation between classes. The key components of the content encoder is using 2 down convolutional blocks (with instance normalisation, and ReLU activation), followed by 4 basic ResNet blocks (with instance normalisation, and ReLU activation), and finally a Gaussian noise layer. The basic ResNet blocks aids the encoding of a class-irrelevant space, and the Gaussian layer prevents the space from becoming zero.

Our generator takes in as input the content and attribute latent spaces. The attribute is first upsampled ($\times 4$, with nearest neighbors) to the same size as the content latent space, concatenated, and then combined using several basic ResNet blocks. Finally, we use deconvolutional blocks (transpose convolution with kernel size of 4, followed by average pooling, layer normalisation \cite{ba2016layer}, and a ReLU activation) to upsample to the original input size.

In addition, not shown in Fig.~\ref{fig:network_architecture}, our domain discriminator contains 6 convolutional layers with leaky ReLUs (kernel size = 3, stride = 2), followed by 2 additional convolutional layers (kernel size = 1, stride = 1), and adaptive average pooling for each class output, and real/ fake output. Our content discriminator contains 3 convolutional layers with leaky ReLUs (kernel size = 3, stride = 2), followed by an additional convolutional layer (kernel size = 4, stride = 1), adaptive average pooling, and a final fully connected layer for class output.

\subsection{Rejection sampling of the attribute latent space.} 

\begin{figure}[!ht]
  \centering
\makebox[\linewidth]{
	\includegraphics[width=0.6\textwidth]{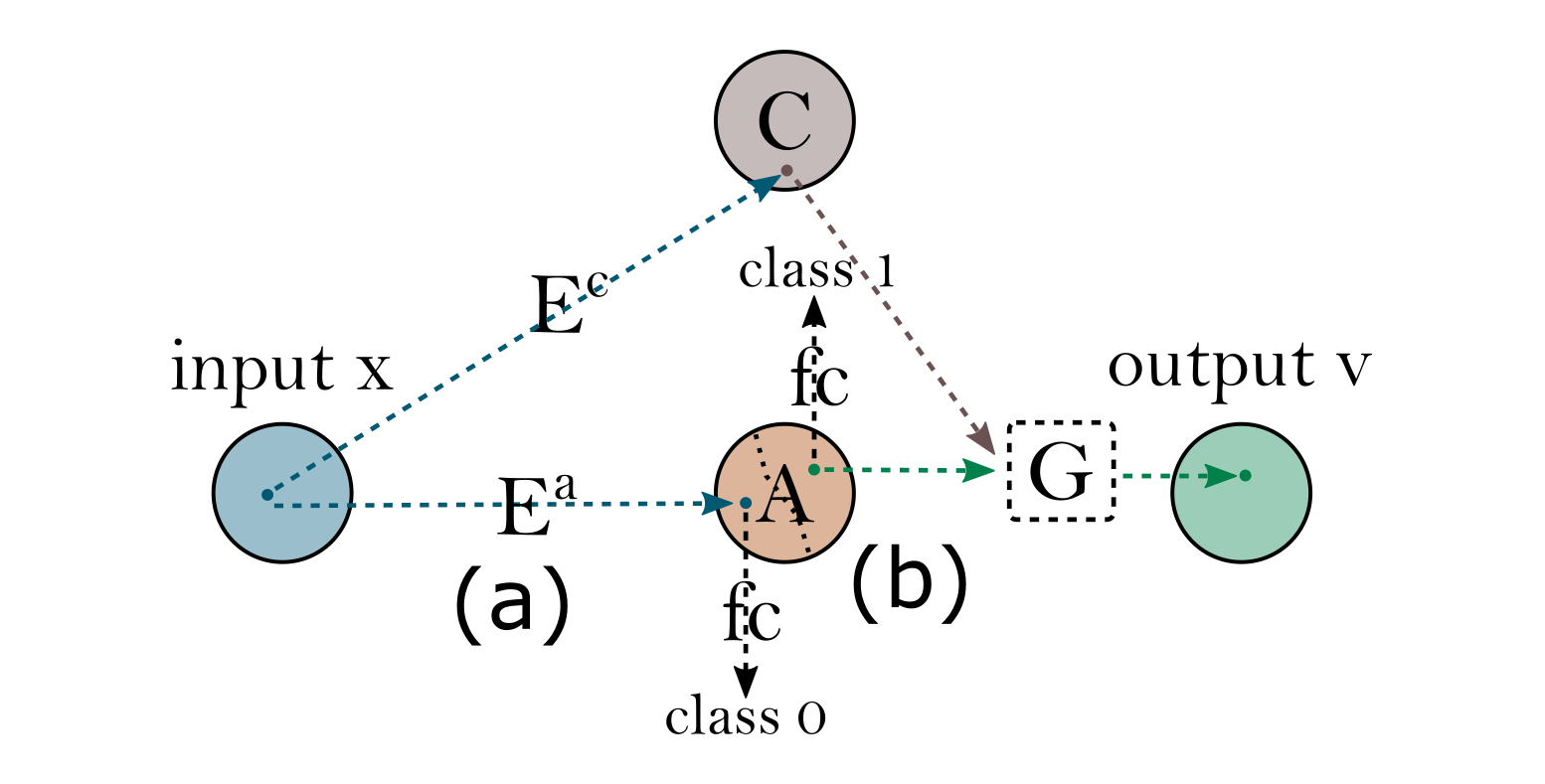}}
\caption{Rejection sampling during training/ testing. Using ICAM, translation can be achieved using a single input image, in addition to translating between 2 images. (a) An input image is encoded into content and attribute spaces, and is passed through the classifier to identify its class (0 in this example). (b) Attribute space A is then randomly sampled until a random vector of the opposite class is sampled (class 1 in this case), by checking its class using the classifier. The newly sampled vector is passed to the generator along with the encoded content space to achieve translation between class 0 and 1.}
\label{fig:random_sampling}
\end{figure}

To further encourage disentanglement of the attribute latent space, during training, since we have a classification layer on the attribute latent space, we are able to control the random sampling of the attribute latent space by checking which class it belongs to (Fig.~\ref{fig:random_sampling}). If a sample is of the wrong class, it is rejected, and another vector is sampled. This allows us to sample a random attribute vector of an appropriate class, thus aiding the quality of random generation, and providing a better signal for the generator during training. Rejection sampling is not possible in a network like DRIT \cite{lee2019drit}, as there is not a classification layer that is applied to the attribute latent space. 


\subsection{The Human Connectome Project (HCP) dataset} 
\label{sec:HCP}

Imaging phenotype variability is common in many neurological and psychiatric disorders, and is an important feature for diagnosis. This type of variation was simulated in \citet{baumgartner2018visual} for a simple 2D case where class 0 was simulated with no features and class 1 was simulated as two sub-types, with one feature in common and then a second feature, which appears in a different location in each sub-population. 

\begin{figure}[!ht]
  \centering
\makebox[\linewidth]{
	\includegraphics[width=0.8\textwidth]{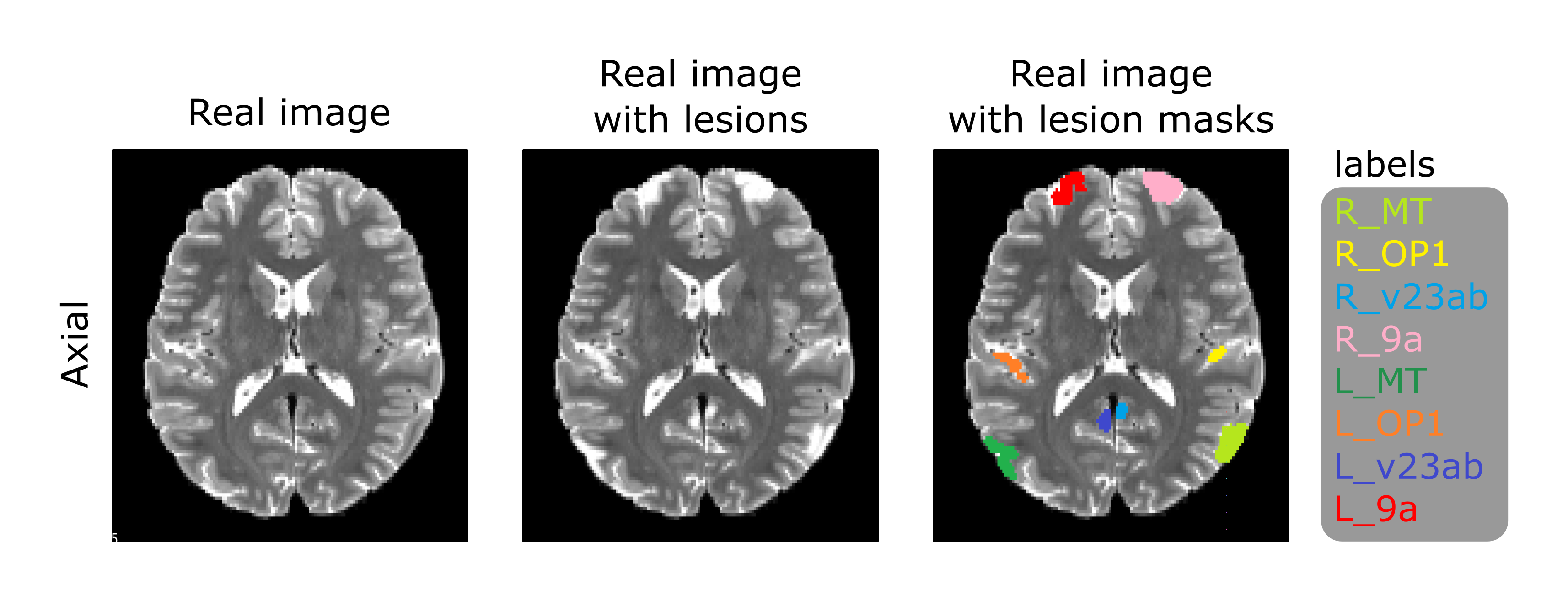}}
\caption{Example of a 2D MRI axial slice from the HCP dataset with and without lesions. We note that the simulated lesions are of similar pixel intensities to the CSF. This is often observed in pathological lesions, and can make them challenging to detect.}
\label{fig:hcp_dataset}
\end{figure}

We use the HCP dataset with T2 MRI volumes to take this simulation further, and to create cortical features or 'lesions' which appear in the 'disease' class with different frequencies (Fig.~\ref{fig:hcp_dataset}). Specifically, eight regions were selected from the HCP parcellation \cite{glasser2016multi}. These were defined as binary masks for cortical surface meshes and then mapped back to the cortical volume (with a 2mm thickness), using HCP 'workbench' tools \cite{marcus2011}, followed by intensity scaling to match intensity of the cortical spinal fluid, and finally by a Gaussian blur filter. 

Every simulated example includes common regions with a further locations were selected as having lesion or no lesion using a random number generator. The regions selected are; $MT_R$, $MT_L$ (medial temporal area), $OP1_R$, $OP1_L$ (posterior opercular cortex), $v23ab_R$, $v23ab_L$ (posterior cingulate cortex), $9a_R$, $9a_L$ (medial prefrontal cortex), where R and L are right and left, respectively. $MT_R$ and $MT_L$ were selected as common regions, and appear in every subject. To be able to compare against DRIT, 2D networks were trained on 2D axial slices from the centre of the brain (to which all regions are constrained by design). Since there is a lot of variability between subjects, not all the lesions appear in the selected slice for each subject, and so there is further variability in lesion appearance.

Prior to training, the T2 images were bias corrected, brain extracted and linearly aligned (for full details on image acquisition and pre-processing see \cite{glasser2013minimal}). Images were normalised in range [0, 1] per subject, and resized to $128\times160\times128$ voxels, and sliced into 2D axial images, of size $128\times160$. Data was randomly split for training, validation and testing using an 80/10/10 ratio (712, 88, 88 subjects each, respectively), consistently for all networks.

\subsection{Alzheimers Disease Neuroimaging Initiative (ADNI) dataset} 
\label{sec:ADNI}

Used also in \citet{baumgartner2018visual} the longitudinal ADNI dataset \cite{jack2008alzheimer} supports ground-truth evaluation of the feature attribution maps as it contains a subset of paired examples for which images are acquired before and after conversion to full a AD state; where the intermediate state between healthy cognition and full dementia is known as mild cognitive impairment (MCI). The 3T acquired T1 MRI volumes were N4 bias corrected \cite{tustison2010n4itk} using simpleitk, brain extracted using freesurfer \cite{segonne2004hybrid} and rigidly registered to the MNI space using Niftyreg \cite{modat2012inverse}. Images were normalised in range $[-1, 1]$ per subject, and resized to $128\times160\times128$ voxels. We split the dataset into AD and MCI classes, with 257 and 674 volumes used in training, respectively. For testing, 61 subjects which convert from MCI to AD (i.e. paired subjects) are used. A further 61 conversion subjects are used for validation. To compute the disease maps, these paired subjects were in addition rigidly aligned to each other, and the difference between them was the disease map for that pair. Finally all disease and FA maps were masked to ensure that the returned NCC values reference brain tissue only.

\subsection{UK Biobank dataset} 
\label{sec:biobank}

UK Biobank data included \cite{alfaro2018image, miller2016multimodal} 11,735 T1 3D MRI volumes, selected from two age bins 45-60 years (class 0, on average $54.6 \pm 3.4$ years) and 70-80 years (class 1, on average $73.0 \pm 2.2$ years). T1 image processing (see also \cite{alfaro2018image}) involved bias correction using FAST \cite{zhang2001segmentation}, brain extraction using BET \cite{smith2002fast} and linear registration to MNI space, using the FLIRT toolbox \cite{jenkinson2002improved}. Our young subjects are separated into training, validation, and testing sets with 6706, 373 and 372 in each, respectively. Our older subjects are separated into training, validation, and testing sets with 3856, 214 and 214 in each, respectively. The input into the networks is resized to $128\times160\times128$ voxels, and normalised in range $[0,1]$, per subject. 


\subsection{Comparison methods} \label{sec:comparison_methods}
We compare our proposed approach against a range of baselines in our experiments. For a fair comparison, we train and test all methods on the same training, validation and testing datasets.

\paragraph{Grad-CAM \cite{selvaraju2017grad}, integrated gradients \cite{sundararajan2017axiomatic} and occlusion \cite{zeiler2014visualizing}.}
We trained a simple 3D ResNet with 4 down ResNet blocks, and a fully connected layer for classification. We then used the captum library \cite{captum2019github} implementation of Grad-CAM, integrated gradients and occlusion to generate the feature attribution maps for each method.  

Grad-CAM is gradient-based saliency method that computes the gradients of the target output with respect to the final convolutional layer of a network. The layer activations are weighted by the average gradient for each output channel and the results are summed over all channels to produce a coarse heatmap of important for prediction of a class. 

Integrated gradients is another method of analysing the gradient of the prediction output with respect to features of the input. It is defined as the integral of the gradients along the straight line path from a given baseline to the input image. A series of images are interpolated between the baseline (e.g. matrix of 0s) and the original image, and the integrated gradients are given by the integration of the computed gradients for all the images in the series.

Occlusion is a perturbation-based method that involves replacing portions of an image with a block of a given baseline value (e.g. 0), and computing the difference in output. A heatmap is formed using the difference between the output probability attributed to the original volume and the probability computed for the occluded volume, for different positions of the occlusion block across the input image.

Grad-CAM was implemented on the last convolutional block of the ResNet, with a size of $4\times5\times4$, and was up-sampled to the input size for visualization. 
For the implementation of integrated gradients we considered a baseline volume with constant value of 0, and the integral was computed using 200 steps.
Occlusion was implemented using occlusion blocks with value 0, size $10\times10\times10$ and stride 5. 

\paragraph{DRIT \cite{lee2019drit}.} In our ablation experiments, we use the 2D network DRIT++ described in \citet{lee2019drit}. Because we aim to use the network for feature attribution, instead of its original goal of domain translation, we had to make some changes to the original network. The aim of the ablation experiments were to assess the different components of ICAM, so we used the most comparable version of DRIT. In particular, we use the DRIT++ network, which has a shared generator, but use an unconditional version (i.e. without the input of class label) of the network, so that it is comparable to ICAM. We also generate the feature attribution map in the same way.

\paragraph{VA-GAN \cite{baumgartner2018visual}.}
We used the VA-GAN network for feature attribution, as described in the original paper.

\paragraph{Model selection.} 
For VA-GAN, $ICAM_{DRIT}$ and $ICAM$, the last model is selected in the Biobank experiments. In all other experiments, the models selected are based on the best model result on the validation dataset, using the NCC score. For Grad-CAM, integrated gradients and occlusion, as the FA maps are only generated after a network is trained, we could not select a model based on its performance with the NCC score, during training/ validation. We instead selected the best model based on the accuracy classification score on the validation dataset, to prevent the effect of overfitting. 

\subsection{Training details} \label{sec:training}

We used PyTorch \cite{Paszke2019PyTorch} Python package in all of our deep learning experiments, and trained using NVIDIA TITAN GPUs. We trained our networks in a similar fashion to \citet{lee2019drit}. During training in each iteration, the content discriminator is updated twice, followed by the update of the encoders, generators, and domain discriminators (i.e. each training iteration uses 3 batches to perform these updates). For each update of the generator, an input is selected for each class (e.g. 2 inputs including class 0 and 1) to achieve translation. In addition, each input is encoded and translated to the opposite class by randomly sampling the attribute latent space, and obtaining an appropriate class, using the classifier. 

In all experiments, unless otherwise stated, we used the following hyperparameters during training of ICAM networks: learning rate for content discriminator = 0.00004, learning rate for the rest = 0.0001, Adam optimiser with betas = (0.5, 0.999), $\lambda_{D^c}=1, \lambda_{D}=1, \lambda_{BCE}=10, \lambda_{KL}=0.01, \lambda_{M}=10, \lambda_{z^a}=1, \lambda_{rec}=100, \lambda_{D_{BCE}}=1$ for discriminator optimisation, and $\lambda_{D_{BCE}}=5$ for generator optimisation.

In the UK Biobank experiments, we trained all networks for 50 epochs. In the HCP 2D ablation and ADNI experiments, all networks (including VA-GAN and DRIT) were trained for 300 epochs. In the ADNI experiments, because we had a limited dataset, we further refined ICAM with updated lambdas ($\lambda_{rec}=10$, and $\lambda_{BCE}=20$) for another 200 epochs. We could not refine VA-GAN any further because generator and discriminator losses went to zero during training, often after 150 epochs.

\paragraph{Baseline methods.} For training VA-GAN, and DRIT, we used the default parameters as provided in the original papers and publicly released code repositories. For Grad-CAM, integrated gradients, and occlusion, the classifier network was trained with learning rate of 0.0001, SGD with momentum of 0.9, for 50 epochs, and using a weighted BCE loss to account for class-unbalanced training data. Since the model converged by 50 epochs, we did not train for any longer. 

\section{Results} 

\subsection{HCP experiments} 

\begin{figure}[!p]
  \centering
\makebox[\linewidth]{
	\includegraphics[width=1.0\textwidth]{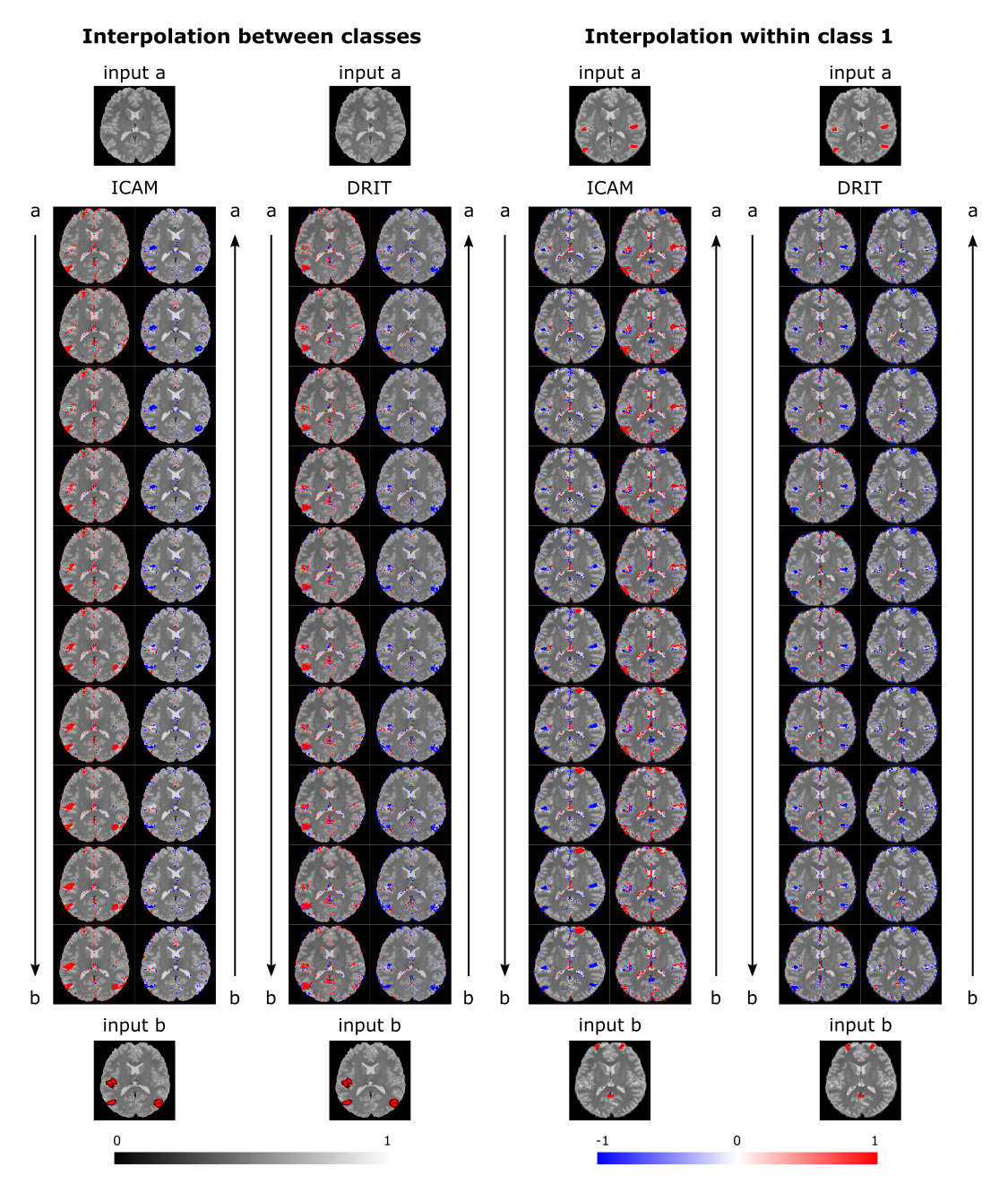}}
\caption{Interpolation between classes 0 (no lesions) and 1 (lesions), and within class 1. Red, addition of lesions; blue, removal of lesions.}
\label{fig:interpolation}
\end{figure}

In our additional HCP experiments (see section \ref{sec:HCP} for dataset details), we show that ICAM can interpolate between classes (Fig.~\ref{fig:interpolation}, left), showing a smooth interpolated result, as well as capturing the large majority of the lesions in both addition and removal, whereas DRIT does not visibly demonstrate smooth interpolation, and is only able to do removal of lesions effectively.
This suggests that the content latent space has not been completely disentangled, and that it might encode some class information (about the lesions). That is most likely caused by the network architecture, as encoding the attribute latent space as a 1D vector, instead of a 2D vector as with ICAM, might not be sufficient to encode spatial information about the lesions. In addition, ICAM shows visibly better interpolation within class 1 in the latent space (Fig.~\ref{fig:interpolation}, right). ICAM is able to both add and remove lesions simultaneously during interpolation, while DRIT is only able to remove lesions.

\begin{figure}[!ht]
  \centering
\makebox[\linewidth]{
	\includegraphics[width=1.0\textwidth]{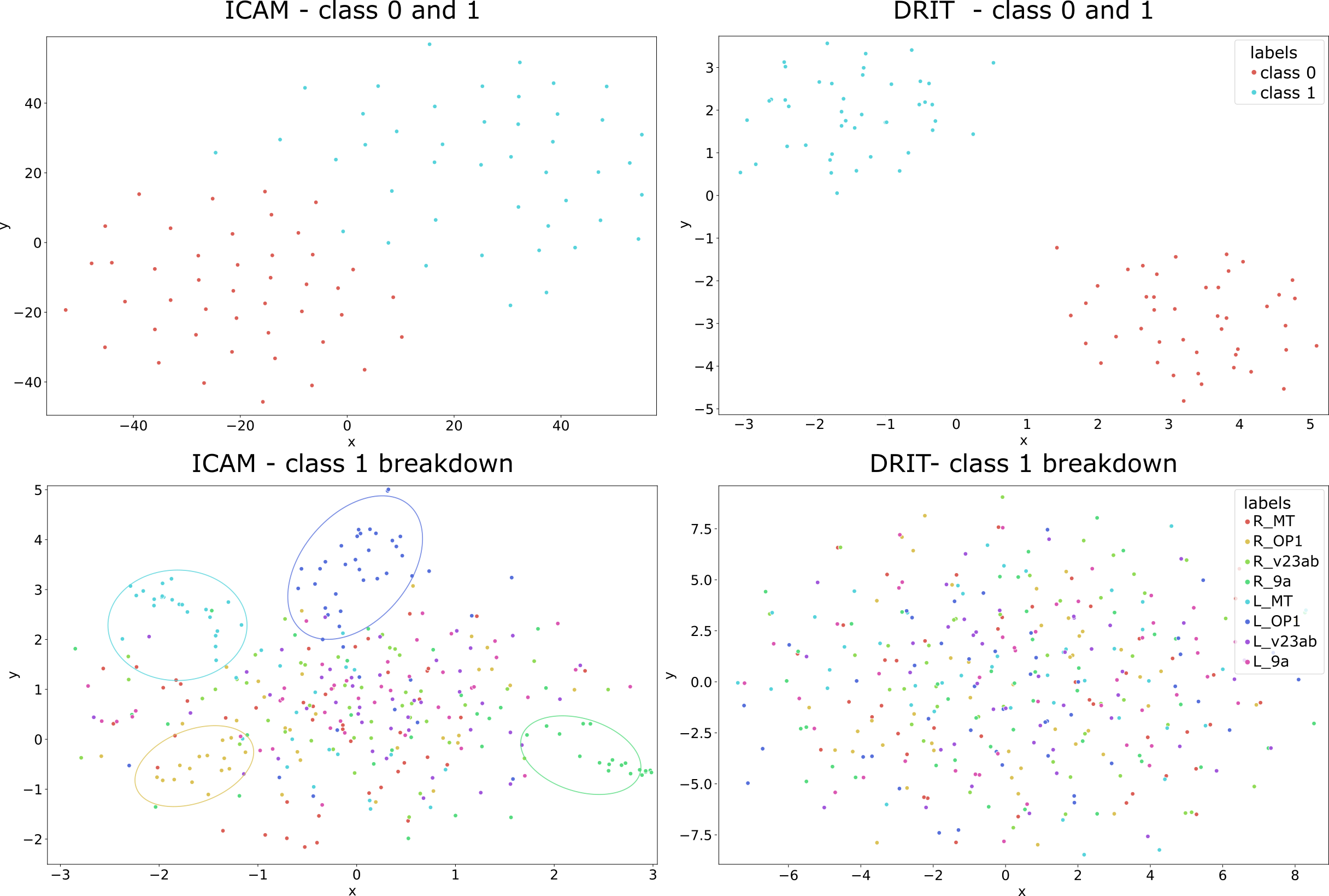}}
\caption{tSNE plots for ICAM and DRIT methods using the HCP test dataset.}
\label{fig:tsne}
\end{figure}

Finally, ICAM demonstrates clear separation between class 0 (no lesions) and 1 (lesions) in its latent space (Fig.~\ref{fig:tsne}, top, left). In addition, without explicitly giving information about the different lesions, ICAM is able to cluster several of the lesions in the latent space (shown via the circles), when encoding tSNE for class 1 (lesions with 8 different locations) examples (Fig.~\ref{fig:tsne}, bottom, left). DRIT is also very separable between classes 0 and 1 (Fig.~\ref{fig:tsne}, top, right), but does not demonstrate separation within class 1 (Fig.~\ref{fig:tsne}, bottom, right). 

\subsection{ADNI experiments}
\begin{figure}[!p]
  \centering
\makebox[\linewidth]{
	\includegraphics[width=1.0\textwidth]{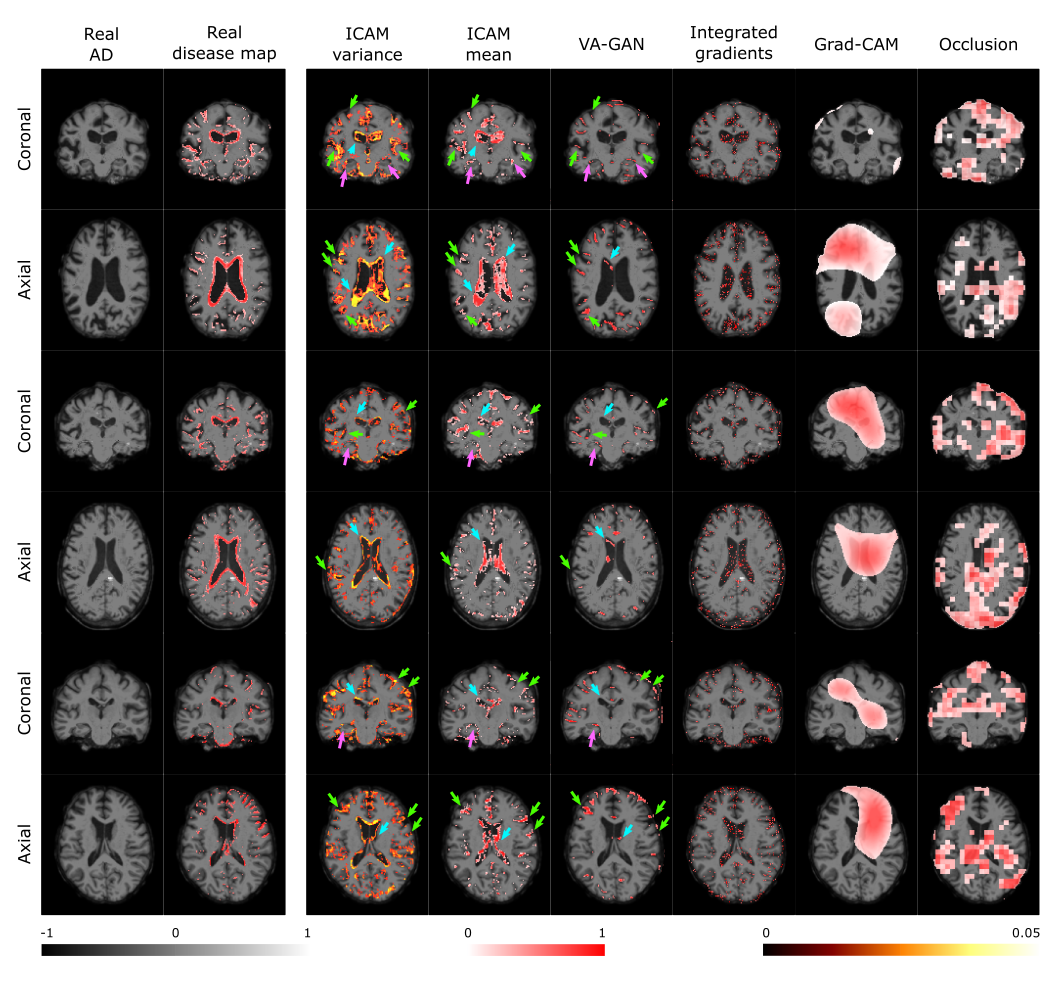}}
\caption{Additional AD to MCI ADNI comparisons for FA map generation in 3 subjects. Blue arrows, ventricles; green arrows, cortex; pink arrows, hippocampus.}
\label{fig:adni_extra_comparisons}
\end{figure}

We show additional examples (see section \ref{sec:ADNI} for dataset details) of comparisons between ICAM and baseline methods in Fig.~\ref{fig:adni_extra_comparisons}. In general,  we observe ICAM achieves visibly better detection compared to baseline methods, with the variance map of ICAM most sensitive to variability in the cortex (green arrows), ventricles (blue arrows), and hippocampus (pink arrows), and appeared to be the most similar to the real disease maps (e.g. rows 3-4). 

We note that ICAM and VA-GAN seem to detect some differences which do not appear in the ground truth disease map (e.g. hippocampal atrophy, pink arrows, row 5). Some of this may relate to measurement noise, for example motion is known to be a significant challenge in dementia datasets. Overall ICAM variance maps flag up more evidence of disease than VA-GAN; however it is important to stress that, MCI-AD conversion is not a binary process. These maps are therefore likely picking up heterogeneity in the relative timing and disease progression in these subjects.


\subsection{Biobank experiments}
\begin{figure}[!p]
  \centering
\makebox[\linewidth]{
	\includegraphics[width=1.0\textwidth]{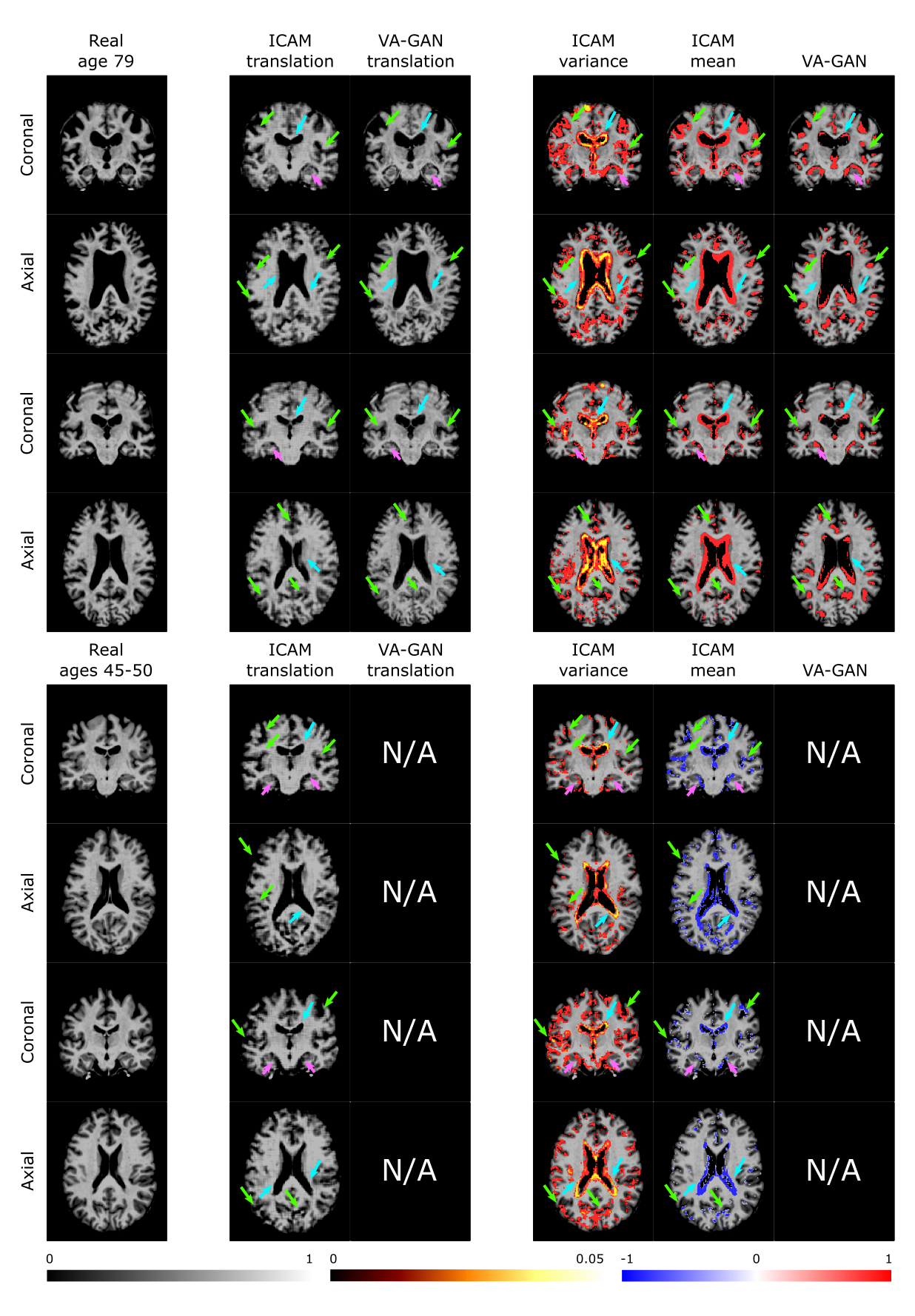}}
\caption{Biobank translation results for translation of old to young (top) and translation of young to old (bottom) in 4 subjects. Blue arrows, ventricles; green arrows, cortex; pink arrows, hippocampus.}
\label{fig:biobank_very_old_young}
\end{figure}

In our additional Biobank experiments (see section \ref{sec:biobank} for dataset details), we performed translation with very old ($>79$ years), and young ($<50$ years) subjects (Fig.~\ref{fig:biobank_very_old_young}). VA-GAN is not a cyclic network and therefore does not have an example for young to old translation. We found very high detection in relevant brain regions for both ICAM and VA-GAN for translation of old subjects to young. VA-GAN appeared to detect more hippocampal differences than ICAM when comparing to ICAM's FA mean maps, but appeared similar when comparing to the variance maps (rows 1 and 3). In addition, ICAM was able to detect much higher variability in the ventricles and was able to change the shape of the brain, while VA-GAN was only able to make minor adjustments in pixel intensities. While we found weaker detection in our young to old translation (Fig.~\ref{fig:biobank_very_old_young}, bottom), many of the expected regions were still highlighted in the FA maps, and we even observed changes in brain shape, including enlargement of the ventricles (blue arrows) and decrease in cortical thickness (green arrows).

\begin{figure}[!p]
  \centering
\makebox[\linewidth]{
	\includegraphics[width=1.0\textwidth]{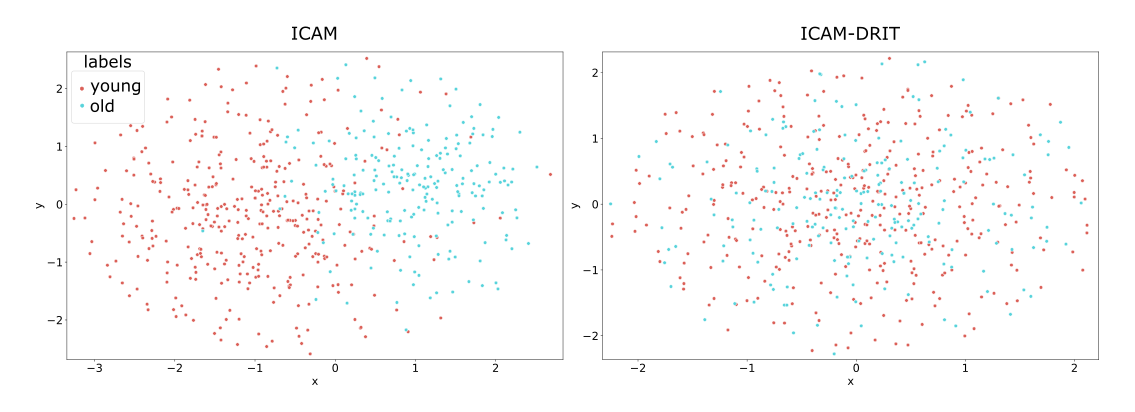}}
\caption{tSNE plots for Biobank on age classification.}
\label{fig:tsne_biobank}
\end{figure}

\begin{figure}[!p]
  \centering
\makebox[\linewidth]{
	\includegraphics[width=0.7\textwidth]{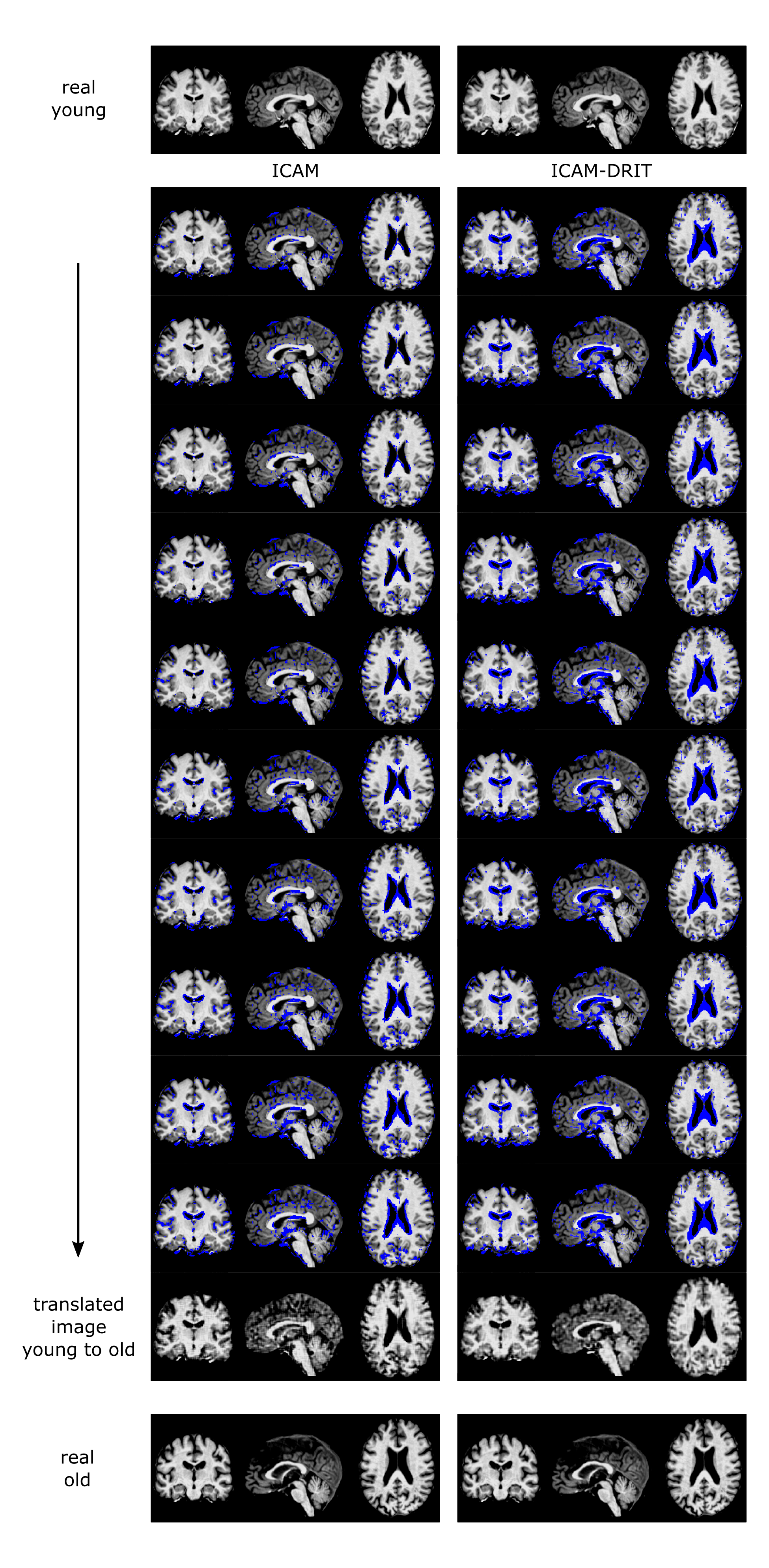}}
\caption{Biobank interpolation between class 0 (young) and 1 (old) for $ICAM$ and $ICAM_{DRIT}$.}
\label{fig:biobank_interpolation}
\end{figure}

To demonstrate the effectiveness of the added components in $ICAM$, we compared against a baseline version of our network, $ICAM_{DRIT}$, as DRIT cannot scale to 3D. We performed interpolation between old and young subjects (Fig.~\ref{fig:biobank_interpolation}), and found that with $ICAM_{DRIT}$, the FA maps are similar across the interpolation. It is possible that the attribute and content spaces have not been disentangled, and that some class relevant information was encoded in the content space. In contrast, $ICAM$ demonstrates smooth interpolation between the classes, and thus is likely to have disentangled age in the attribute latent space. Additional evidence for this is the tSNE plots of $ICAM$, and $ICAM_{DRIT}$ (Fig.~\ref{fig:tsne_biobank}), where we see separation between old and young subjects for $ICAM$, but no separation for $ICAM_{DRIT}$.

\end{document}